\RequirePackage{iftex}
\ifPDFTeX
  \pdfoutput=1
\fi
\documentclass{article}

\usepackage{microtype}
\usepackage{graphicx}
\usepackage{subcaption}
\usepackage{booktabs} 
\usepackage[table,dvipsnames]{xcolor}
\usepackage{adjustbox}
\usepackage{comment}
\usepackage{amsmath}
\usepackage{amssymb}

\usepackage{tcolorbox}
\tcbuselibrary{listings, breakable, skins}

\newtcolorbox{promptbox}[1]{
  colback=gray!5,
  colframe=black,
  fonttitle=\bfseries,
  title=#1,
  breakable,
  enhanced,
  boxrule=0.8pt,
  arc=2pt,
  left=6pt,
  right=6pt,
  top=6pt,
  bottom=6pt
}

\usepackage{setspace}

\usepackage{ragged2e}

\usepackage{listings}
\usepackage{xcolor}
\usepackage{hyperref}

\usepackage[accepted]{icml2026}

\usepackage{amsmath}
\usepackage{amssymb}
\usepackage{mathtools}
\usepackage{amsthm}
\usepackage{multirow}

\definecolor{LightLav}{HTML}{F3ECFF}

\usepackage[capitalize,noabbrev]{cleveref}

\theoremstyle{plain}

\theoremstyle{definition}

\theoremstyle{remark}

\usepackage[textsize=tiny]{todonotes}

\icmltitlerunning{DREAM-R: Multimodal Speculative Reasoning with RL-Based Refined Drafting, Precise Verification, and Fully Parallelism}

\begin{document}

\twocolumn[
  \icmltitle{DREAM-R: Multimodal Speculative Reasoning with RL-Based Refined Drafting, Precise Verification, and Fully Parallel Execution}



  \icmlsetsymbol{equal}{*}

  \begin{icmlauthorlist}
    \icmlauthor{Yunhai Hu}{nyu}
    \icmlauthor{Zining Liu}{upenn}
    \icmlauthor{Xiangyang Yin}{nyu}
    \icmlauthor{Tianhua Xia}{nyu}
    \\\icmlauthor{Bo Bao}{crbs}
    \icmlauthor{Eric Sather}{crbs}
    \icmlauthor{Vithursan Thangarasa}{crbs}
    \icmlauthor{Sai Qian Zhang}{nyu}
  \end{icmlauthorlist}

  \icmlaffiliation{nyu}{New York University}
  \icmlaffiliation{upenn}{University of Pennsylvania}
  \icmlaffiliation{crbs}{Cerebras Systems}

  \icmlcorrespondingauthor{Yunhai Hu}{yh5961@nyu.edu}

  \icmlkeywords{Machine Learning, ICML}

  \vskip 0.3in
]



\printAffiliationsAndNotice{}  

\begin{abstract}
Speculative reasoning has recently been proposed as a means to accelerate reasoning-intensive generation in large multimodal models, but its effectiveness is often constrained by misalignment between speculative drafts and target-verified reasoning. In this work, we introduce \textit{DREAM-R}, a framework that substantially improves the performance of speculative reasoning. At its core, DREAM-R employs \textit{Speculative Alignment Policy Optimization} (SAPO), a reinforcement-learning objective that trains draft models to generate reasoning steps that are both faithful to target trajectories and concise. We further propose a \textit{Contrastive Probability Normalization} (CPN) that uses a ratio-based criterion to provide stable and interpretable acceptance of speculative steps only when positive evidence clearly dominates, thereby preventing error propagation. Building on these components, we develop a \textit{Fully Parallel Speculative Reasoning} (FPSR) framework that parallelizes draft generation, target-side reasoning, and verification across multi-step reasoning, enabling early stopping and clean fallback. Experiments on reasoning-heavy benchmarks demonstrate up to $2.48\times$ speedup while preserving target-model accuracy, yielding substantial efficiency gains without compromising reasoning quality. Code is available at \url{https://github.com/HuYunhai-Alex/DREAM-R}.

\end{abstract}

\section{Introduction}
Large models (LMs)~\cite{grattafiori2024llama,brown2020language,ouyang2022training,achiam2023gpt} have become the backbone of modern artificial intelligence, driving major advances across language understanding, vision, and multimodal perception by leveraging massive model capacity and data scale. While these models excel at representation learning and pattern recognition, many real-world tasks require more than direct prediction, demanding structured, multi-step inference and long-horizon decision-making. This need has given rise to large reasoning models (LRMs)~\cite{bai2025qwen2,hong2025glm,wang2025internvl3}, which extend large models with explicit reasoning capabilities that can decompose complex problems, integrate heterogeneous evidence, and maintain consistency across extended inference chains. LRMs are particularly important in settings such as multimodal understanding, scientific problem solving, and embodied intelligence, where accurate outcomes depend on coherent reasoning rather than surface-level correlations. However, the increased computational cost of reasoning-intensive generation presents significant efficiency challenges, motivating research into methods that can accelerate reasoning while preserving accuracy.

To address this challenge, recent advances in speculative decoding~\cite{li2024eagle,li2024eagle2, li2025eagle3, cai2024medusa, ankner2024hydra, xia2023speculative, zhang2023draft, miao2023specinfer, chen2024sequoia, sun2024triforce,hu2025speculative,hu2025dream} have shown promise in accelerating inference by parallelizing generation and verification using lightweight draft models. Building on this idea, speculative reasoning~\cite{pan2025specreason,fu2025scaling} extends speculation beyond token-level decoding to reasoning-step granularity, enabling more effective acceleration for LRMs by exploiting parallelism while preserving correctness. These developments point toward a promising direction for scaling reasoning performance without proportionally increasing inference cost.

Most existing speculative decoding and speculative reasoning methods are developed and evaluated primarily for large language models (LLMs) and LRMs, with limited investigation into multimodal LLMs (MLLMs), especially multimodal large reasoning models (MLRMs). In practice, we observe that directly applying these techniques to MLRMs often yields suboptimal performance, as the reasoning characteristics and error modes in multimodal settings differ substantially from those of text-only models. Through an empirical analysis of multimodal reasoning behavior, we observe that design choices effective for LLMs do not transfer reliably to MLRMs. Motivated by these observations, we introduce~\textit{DREAM-R}, a multimodal speculative reasoning framework that integrates RL-based refined drafting, precise threshold-based verification, and fully parallel execution to effectively accelerate multimodal reasoning while preserving accuracy. In particular, our contribution can be summarized as follows:
\begin{itemize}
\item DREAM-R incorporates a ~\textit{Speculative Alignment Policy Optimization} (SAPO), a reinforcement-learning objective that trains draft models to produce reasoning steps that are both faithful to target-model trajectories and concise.
\item Furthermore, we introduce a~\textit{Contrastive Probability Normalization} (CPN) that employs a contrastive probability normalization to ensure stable and interpretable acceptance of speculative steps only when positive evidence clearly outweighs negative evidence, effectively preventing error propagation.
\item Finally, we present a~\textit{Fully Parallel Speculative Reasoning} (FPSR) framework that parallelizes draft generation, target-side reasoning, and verification across multi-step reasoning, enabling early stopping and clean fallback.
\item Experiments on reasoning-heavy benchmarks demonstrate up to $2.48\times$ speedup while maintaining target-model accuracy, delivering substantial efficiency gains without compromising reasoning quality.
\end{itemize}

\section{Related Works}


\paragraph{Large Reasoning Models} Large reasoning models (LRMs) are distinguished by their ability to generate extended chains of thought (CoT)~\cite{wei2022chain}, enabling explicit intermediate reasoning steps prior to producing a final answer. In language-only settings, this inference paradigm has yielded substantial gains on complex tasks such as mathematical reasoning~\cite{cobbe2021training} and code generation~\cite{chen2021evaluating}. For vision-language reasoning models (VLRMs), CoT reasoning naturally extends to multimodal scenarios that require integrating visual evidence with textual inference. Recent VLRMs~\cite{bai2025qwen2,hong2025glm,wang2025internvl3} demonstrate strong performance on vision-dependent reasoning benchmarks, highlighting the effectiveness of multimodal CoT reasoning. 

\begin{figure}
     \centering
     \includegraphics[width=\columnwidth]{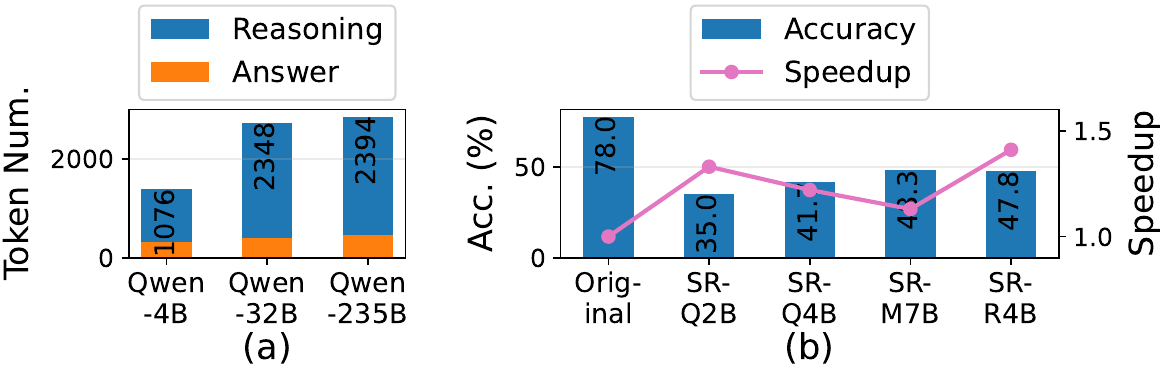}
     \caption{(a) Numbers of reasoning and answer tokens. Qwen-4B, Qwen-32B, and Qwen-235B refer to Qwen3-VL-4B, Qwen3-VL-32B, and Qwen3-VL-235B-A22B.
     (b) Accuracy and speedup of Qwen3-VL-32B under different decoding methods. Original denotes standard decoding. SR-Q2B, SR-Q4B, SR-M7B, and SR-R4B denote SpecReason~\cite{pan2025specreason} using Qwen3-VL-2B, Qwen3-VL-4B, MiMo-VL-7B-RL, and Qwen3-VL-R1-VL-4B as draft models, respectively. Speedup is normalized to Original.}
     \label{fig:rw}
\end{figure}

However, VLRMs typically incur high decoding latency, which is largely driven by the generation of intermediate reasoning steps. Analyzing Qwen3-VL-4B, Qwen3-VL-32B, and Qwen3-VL-235B on 50 samples from the MathVerse dataset, we find that the models generate an average of 2,330 tokens per sample, with 1,939 tokens corresponding to intermediate reasoning, accounting for approximately 83\% of the total output. In contrast, answer tokens make up only 17\% of the output, as illustrated in Figure~\ref{fig:rw}(a). These results highlight that the reasoning phase dominates decoding cost and constitutes the primary bottleneck for end-to-end inference latency in VLRMs.

\paragraph{Speculative Decoding}
Speculative decoding accelerates autoregressive inference by pairing a lightweight draft model with a larger target model~\citep{stern2018blockwise}. After the target model completes prefilling, the draft model proposes multiple speculative tokens, which are verified in parallel by the target model using causal attention. If all speculative tokens are accepted, decoding advances without additional target computation; otherwise, decoding resumes from the first rejected token.

Prior work has explored speculative decoding along multiple axes~\citep{xia2023speculative, hu2025speculative}. Algorithmically, some methods replace the draft model with retrieval-based~\citep{yang2023inference, he2023rest} or n-gram-based generators~\citep{ou2024lossless, stewart2024n}. Architecturally, self-speculative approaches reuse shallow target layers or apply layer skipping to generate drafts~\citep{zhang2023draft, xia2024swift, elhoushi2024layer, liu2024kangaroo}, while others introduce lightweight trainable draft heads that share embeddings or output layers with the target model~\citep{li2024eagle, gao2024falcon}. Verification strategies range from sequential acceptance to tree-based parallel verification~\citep{miao2023specinfer, chen2024sequoia}, and recent systems work explores asynchronous or pipelined execution to improve hardware utilization~\citep{chen2023cascade, mcdanel2025pipespec}.

\paragraph{Multimodal Acceleration}
Extending speculative decoding to multimodal models introduces additional challenges, as draft generation must account for both linguistic and visual evidence. Several recent works adapt speculative decoding to multimodal settings by modifying draft model architectures or training strategies. VADUSA~\citep{li2025fast} applies speculative decoding with tolerance mechanisms in text-to-speech systems. For vision-language models, \citet{gagrani2024speculative} pair a language-only draft model with a VLM target, while IbED~\citep{lee2025inbatch} improves efficiency through batch-level ensembling without additional parameters. Other approaches focus on strengthening multimodal draft models: DREAM~\citep{hu2025dream} incorporates cross-attention and adaptive feature selection, ViSpec~\citep{kang2025vispec} compresses visual tokens via lightweight vision adapters, and MASSV~\citep{ganesan2025massv} adapts small language models into multimodal drafters through architectural changes and self-distillation.

\paragraph{Speculative Reasoning} Large reasoning models, including both language-only and multimodal variants, achieve strong performance on complex tasks but incur high inference latency due to extended CoT generation. To alleviate this overhead, recent work proposes speculative reasoning~\cite{pan2025specreason}, where a lightweight draft model generates candidate reasoning steps that are verified by a larger target model. The target model assigns a scalar score to each step to determine acceptance. The key insight is that many intermediate reasoning steps do not require the full capacity of the target model and can be accurately produced by a smaller model, while preserving semantic correctness even when token-level outputs differ.

Building on this line of work, Lookahead Reasoning~\cite{fu2025scaling} proposes an asynchronous architecture that overlaps the drafting and verification stages, allowing the target model to verify earlier steps while new reasoning steps are generated in parallel. However, these approaches are developed and evaluated primarily for language-only models, and our experiments indicate that directly extending them to vision-language models provides limited benefits. To better understand this limitation, we conduct a simple diagnostic study using 1,000 samples from the MathVista dataset, focusing on the behavior of speculative reasoning when there is a large capability gap between the draft and target models. We evaluate the SpecReason~\cite{pan2025specreason} method using Qwen3-VL-32B as the target model and Qwen3-VL-2B, Qwen3-VL-4B, MiMo-VL-7B-RL, and R-4B as draft models, respectively. We compare the accuracy and speedup of SpecReason against the standard decoding baseline. As shown in Figure~\ref{fig:rw} (b), SpecReason's accuracy drops from $78\%$ to $43.2\%$. One contributing factor is perceptual error in vision-language reasoning, where failures stem from incorrect or incomplete visual grounding rather than flawed logic. For example, in the blue tape positioning case, the visual cue exists in the image but is mis-recognized by the draft model, leading to a coherent yet visually unsupported reasoning trace that complicates verification. This highlights that visual reasoning fundamentally requires grounding in images, which standard language-model drafts tend to ignore.

We find that when the target model is a vision-language model, its judgments of reasoning correctness follow different logic and distributions from those of standard language models. As a result, directly using the log-probabilities of its step-level scores as a reliability signal becomes inaccurate and unstable.

\paragraph{Reinforcement-Learning Based Large Model Training}
Reinforcement learning (RL) has emerged as a central component of the post-training pipeline for foundation models, including both LLMs and VLMs~\citep{grattafiori2024llama,bai2025qwen2,liu2024deepseek,guo2025deepseek,comanici2025gemini}. Through alignment, recent models such as DeepSeek-R1~\citep{liu2024deepseek,guo2025deepseek} have demonstrated substantial performance improvements across a broad range of downstream tasks, particularly those involving complex multi-step reasoning. Existing RL-based alignment approaches can be broadly categorized based on the source of their reward signals into two paradigms: reinforcement learning from human feedback (RLHF)~\citep{Ziegler2019FineTuningLM,Stiennon2020LearningTS,ouyang2022training,bai2022constitutional,rafailov2023direct,Sun2023AligningLM,ethayarajh2024kto} and reinforcement learning with verifiable rewards (RLVR)~\citep{Gao2024OnDE,Zeng2025SimpleRLZooIA,Guan2025rStarMathSL,Wang2025ReinforcementLF,Cui2025ProcessRT,Yu2025DAPOAO}. RLHF leverages human preference annotations to guide alignment, whereas RLVR derives reward signals from automated or programmatic verification mechanisms.

Despite its effectiveness, RLHF depends on human preference annotations and primarily captures relative preferences rather than objective correctness. Extensions such as reinforcement learning from AI feedback (RLAIF)~\citep{bai2022constitutional} help reduce annotation costs, but preference-based signals remain limited in their ability to reliably evaluate correctness for reasoning-intensive tasks, including mathematical problem solving and code generation. In contrast, RLVR relies on task-specific verifiers~\citep{cheng2024fullstack,Kydlicek_Math-Verify_Math_Verification} to assess correctness during training, reducing dependence on human feedback and enabling more reliable correctness evaluation.

Following the RLVR paradigm, our work investigates verifier-guided reward design for speculative reasoning, with the goal of improving the quality of draft-generated reasoning steps and, in turn, enhancing both accuracy and latency in speculative reasoning pipelines, as described in Section~\ref{sec:method}.


\section{Method}
\label{sec:method}
In this section, we first present the \textit{Contrastive Probability Normalization} (CPN) for validating drafted reasoning, followed by \textit{Speculative Alignment Policy Optimization} (SAPO), an RL-based alignment method that trains the draft model to produce reasoning steps consistent with multimodal evidence. Finally, we introduce \textit{Fully Parallel Speculative Reasoning} (FPSR), which concurrently performs drafting, target generation, and verification to maximize hardware utilization and reduce wall-clock latency.

\subsection{Overview of DREAM-R}
\label{sec:overview}
\begin{figure}
    \centering
    \includegraphics[width=0.7\columnwidth]{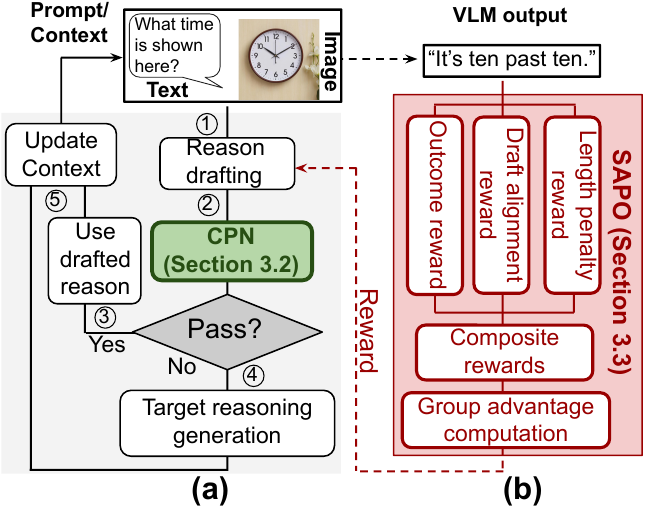}
    \caption{(a) The DREAM-R decoding step. All the step numbers are highlighted in circles. (b) A highlight of SAPO training.}
    \label{fig:overview}
\end{figure}

Figure~\ref{fig:overview} illustrates the overall architecture of DREAM-R. As shown in Figure~\ref{fig:overview}(a), the draft model $M_{\text{draft}}$ first processes the input sample and generates a CoT reasoning trace (Step 1). The target model $M_{\text{target}}$ then performs CPN to assess the quality of this reasoning (Step 2). If the reasoning produced by $M_{\text{draft}}$ passes the verification check (Step 3), it is fed back to $M_{\text{draft}}$ to update the CoT context for the next execution round (Step 5). Otherwise, if the reasoning fails the CPN check (Step 4), the target model $M_{\text{target}}$ directly generates the CoT reasoning from the input.

\subsection{Contrastive Probability Normalization}
\label{sec:tbvm}
\begin{figure}
     \centering
     \includegraphics[width=0.85\columnwidth]{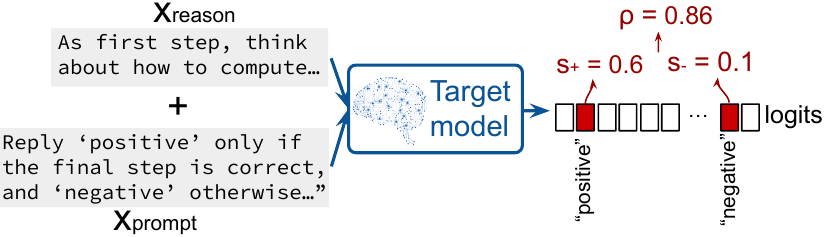}
     \caption{An example on CPN.}
     \label{fig:tbvm}
\end{figure}

As illustrated in Figure~\ref{fig:overview}(a), CPN is used to assess the quality of the reasoning $x_{\text{reason}}$ generated by $M_{\text{draft}}$. DREAM-R adopts a verification scheme in which $x_{\text{reason}}$, together with a predefined prompt template $x_{\text{prompt}}$, is sent to the target model $M_{\text{target}}$. The prompt $x_{\text{prompt}}$ instructs $M_{\text{target}}$ to evaluate the quality of $x_{\text{reason}}$ by selecting one of two keywords: ``positive'' or ``negative''.
As shown in Figure~\ref{fig:tbvm}, let $s_{+}$ and $s_{-}$ denote the predicted probabilities of the keywords ``positive'' and ``negative'', respectively. The verification decision is then made by computing the ratio $\rho = \frac{s_{+}}{s_{+} + s_{-}}$.

This ratio $\rho$ captures the relative dominance between the two signals. When the verifier assigns comparable probabilities to both keywords (e.g., a near 1:1 balance), $\rho$ approaches 0.5, indicating insufficient confidence in a positive judgment. Conversely, larger values of $\rho$ reflect stronger confidence in the positive assessment. Empirically, we adopt a conservative policy: a reasoning step $x_{\text{reason}}$ proposed by $M_{\text{draft}}$ is accepted only if $\rho$ exceeds a predefined threshold $\alpha = 0.7$; otherwise, the reasoning is rejected.

Compared with utility score-based methods such as those proposed in~\cite{pan2025specreason}, which require $M_{\text{target}}$ to assign a discrete score (e.g., from 0 to 9) to the proposed reasoning and accept it if the score exceeds a predefined threshold, CPN provides a more principled and fine-grained assessment. Instead of relying on an arbitrary scalar rating, CPN directly captures the confidence of $M_{\text{target}}$'s judgment by examining the predicted probabilities of the verification keywords. This probabilistic formulation enables a smoother and more interpretable acceptance criterion, reduces sensitivity to prompt-specific scoring biases, and avoids ambiguity introduced by coarse-grained discrete scores. As a result, CPN yields more stable and reliable verification decisions across different reasoning steps and inputs, as demonstrated in Section~\ref{sec:experiment}.


\subsection{Speculative Alignment Policy Optimization}
\label{sec:sapo}

To improve the quality of reasoning produced by $M_{\text{draft}}$, we introduce Speculative Alignment Policy Optimization (SAPO), a policy optimization framework designed for synchronous speculative reasoning. SAPO extends group-based policy optimization by incorporating alignment-aware reward signals that explicitly measure the consistency between the draft model's proposed reasoning steps and the target model's verified reasoning trajectory.

As described in Section~\ref{sec:overview}, during LRM decoding, $M_{\text{draft}}$ iteratively proposes candidate reasoning steps, which are verified by $M_{\text{target}}$ via CPN. This process yields a complete reasoning trace and a final prediction for each input. SAPO assigns a composite verification reward to the resulting reasoning trajectory $x_{\text{reason}}$, consisting of an outcome reward $R_{\text{outcome}}$ that reflects final-answer correctness, a draft alignment reward $R_{\text{draft}}$ that measures the proportion of draft-generated steps accepted by the target model, and a length penalty reward term $R_{\text{length}}$ that penalizes unnecessarily long reasoning traces.
\begin{equation}
R_{\text{outcome}} =
\begin{cases}
1, & \text{if $x_{\text{reason}}$ yields a correct answer},\\
0, & \text{otherwise},
\end{cases}
\end{equation}
while $R_{\text{draft}}$ is defined as $R_{\text{draft}} = \frac{N_{\text{accepted}}}{N_{\text{draft}}}$, where $N_{\text{accepted}}$ and $N_{\text{draft}}$ denote the number of rounds in which the draft reasoning is accepted or rejected by the target model. Finally, to discourage unnecessarily long CoT generation, SAPO employs a length-dependent penalty:
\begin{equation}
\label{eqn:length}
R_{\text{length}} =
\begin{cases}
0, & L \le C,\\
\min\bigl(1,\; k \cdot (L - C)\bigr), & L > C.
\end{cases}
\end{equation}
where $L$ denotes the token length of the reasoning and $C$ is a predefined threshold. Equation~\ref{eqn:length} imposes a penalty proportional to the excess length beyond $C$. The overall reward is computed as
\begin{equation}
\label{eqn:reward-function}
R = w_{1} \cdot R_{\text{outcome}}
  + w_{2} \cdot R_{\text{draft}}
  - w_{3} \cdot R_{\text{length}},
\end{equation}
where $w_{1}$, $w_{2}$, and $w_{3}$ determine the relative contribution of each component. After computing the rewards, SAPO updates the draft model using a simplified Group Relative Policy Optimization (GRPO) objective~\cite{shao2024deepseekmath}, with the group baseline defined as $b_{\mathcal{G}} = \frac{1}{n} \sum_{i=1}^N R_i$,
where $N$ is the number of input samples. $R_i$ is the reward corresponding to the $i$-th draft reason. The group-relative advantage for each trajectory is $A_i = R_i - b_{\mathcal{G}}$.
To improve stability, advantages $A_i$ are normalized within the group, leading to $\tilde{A}_i = \frac{A_i}{\sigma_{\mathcal{G}} + \delta}$,
where $\sigma_{\mathcal{G}}$ denotes the standard deviation of $\{A_i\}$, and $\delta$ is a small constant introduced to improve stability. The policy is updated using a clipped ratio objective:
\begin{equation}
L_i(\theta)
= \min\!\left(
u_i(\theta) \tilde{A}_i,\;
\operatorname{clip}\!\left(
u_i(\theta),\;
1-\epsilon,\;
1+\epsilon
\right) \tilde{A}_i
\right)
\end{equation}
where $\theta$ denotes the parameter set. $u_i(\theta) = \frac{\pi_\theta(a_i \mid x_i)}{\pi_{\theta_{\text{old}}}(a_i \mid x_i)}$ that follows the definition in~\cite{shao2024deepseekmath}, the overall SAPO process is highlighted in Figure~\ref{fig:overview} (b).



\subsection{Fully Parallel Speculative Reasoning}
\label{sec:fpsr}
\begin{figure}
     \centering
     \includegraphics[width=0.82\columnwidth]{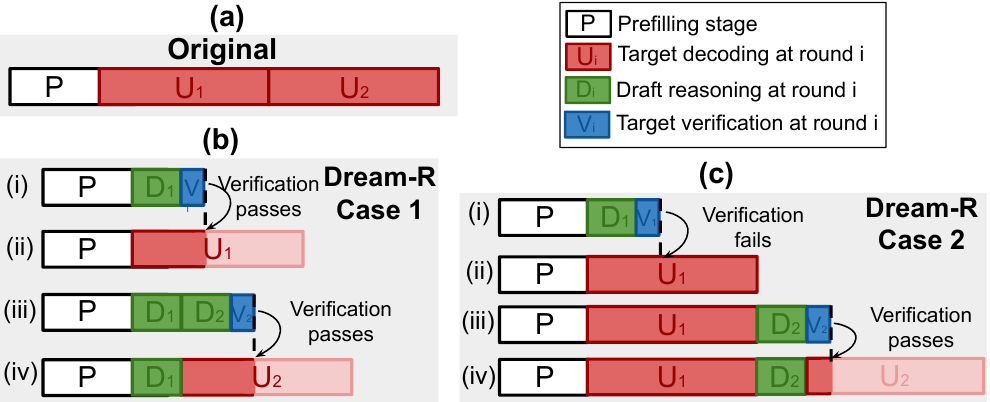}
     \caption{Timeline comparison of (a) Standard Autoregressive Reasoning, and (b) Fully Parallel Speculative Reasoning (FPSR). FPSR pipelines the drafting (Green), verification (Blue), and target generation (Red) stages, ensuring the GPU is never idle. Unlike lookahead methods, we allow 'rollback' (c) to recover from verification failures without re-computation.}
     \label{fig:fpsr}
\end{figure}
While Sections~\ref{sec:tbvm} and~\ref{sec:sapo} focus on the algorithmic design of DREAM-R to improve the reasoning quality of the LRM, this section introduces an advanced Fully Parallel Speculative Reasoning (FPSR) approach that enables high parallelism across the drafting, verification, and decoding stages to further reduce latency.

The FPSR method is illustrated in Figure~\ref{fig:fpsr}. For ease of exposition, we assume a total of two rounds of reasoning. Figure~\ref{fig:fpsr} (a) depicts the baseline approach, in which the target model $M_{\text{target}}$ sequentially performs the prefilling stage P, generates intermediate reasoning steps $U_1$ and produces the final answer $U_2$, resulting in high end-to-end generation latency.

In contrast, Figure~\ref{fig:fpsr} (b) shows the FPSR workflow. At each reasoning round, the draft model $M_{\text{draft}}$ continuously proposes candidate reasoning steps (e.g., $D_1$ in Figure~\ref{fig:fpsr} (b)(i)), while the target model $M_{\text{target}}$ concurrently generates its own reasoning ($U_1$ in Figure~\ref{fig:fpsr} (b)(ii)). Once $M_{\text{draft}}$ completes a reasoning step, it is immediately verified ($V_1$ in Figure~\ref{fig:fpsr} (b)(i)). If the draft reasoning is accepted, $M_{\text{target}}$ early-terminates its ongoing reasoning generation ($U_1$ in Figure~\ref{fig:fpsr} (b)(ii)). This process repeats for subsequent reasoning rounds, assuming the draft reasoning continues to pass verification. As a result, FPSR significantly reduces overall latency compared to the baseline scheme shown in Figure~\ref{fig:fpsr} (a).

In contrast, when the draft reasoning generated by $M_{\text{draft}}$ fails verification (Figure~\ref{fig:fpsr}(c)), the subsequent round cannot begin until the current update stage completes ($U_1$ in Figure~\ref{fig:fpsr} (c)(ii)). Once $U_1$ finishes, a second reasoning round is initiated ($D_2$ in Figure~\ref{fig:fpsr} (c)(iii)) and then verified ($V_2$ in Figure~\ref{fig:fpsr} (c)(iii)). Assuming this verification succeeds, the final result is produced in the $D_2$ (Figure~\ref{fig:fpsr} (c)(iv)).

Compared to prior parallelization approaches that only overlap target generation~\cite{fu2025scaling} or speculative reasoning schemes that perform step verification largely sequentially as shown in Figure~\ref{fig:fpsr} (a), FPSR maintains continuous parallelism across all three components: the draft model proposing candidate reasoning steps, the target model extending the trusted reasoning prefix, and the verifier determining acceptance or termination. By tightly interleaving proposal, verification, and fallback, FPSR significantly reduces end-to-end wall-clock latency for reasoning-intensive decoding, while fully preserving the target model's answer quality.

\subsection{DREAM-R System Implementation}
\label{sec:system-implementation}
We implement a speculative reasoning engine consisting of four tightly coordinated components. The \emph{Drafter} continuously proposes speculative reasoning steps whenever the \texttt{can\_draft} signal is active, and each proposal is forwarded to the \emph{Verifier} for binary acceptance or rejection. If a step is rejected, the \emph{Corrector} is invoked to produce a revised, non-speculative replacement. A centralized \emph{Manager} orchestrates ordered commitment, rollback, and task cancellation across all components, while enforcing a bounded lookahead window to prevent unbounded speculation.

The control flow proceeds as follows. Drafting runs asynchronously until a rejection occurs. When a speculative step is rejected, all in-flight speculative tasks within the lookahead window are canceled, the system rolls back to the most recent verified prefix, and the corrected step is committed in its place. Drafting then resumes from the updated context. This design enables high-throughput speculative reasoning while strictly preserving logical consistency and stepwise correctness. Detailed prompts are provided in the Appendix~\ref{sec:prompts} for completeness.

\section{Experiments}
\label{sec:experiment}
We evaluate our fully parallel speculative reasoning framework on four well-established multimodal reasoning benchmarks, MathVerse~\cite{zhang2024mathverse}, MMBench~\cite{liu2024mmbench}, RealWorldQA~\cite{realworldqa2024}, and MMMU~\cite{yue2024mmmu}, using their full evaluation splits. We experiment with multiple draft and target model combinations. The draft model is chosen from  Qwen3-VL-2B (Q2B)~\cite{yang2025qwen3}, Qwen3-VL-4B (Q4B), R-4B (R4B)~\cite{yang2025r}, and MiMo-VL-7B-RL (M7B-RL)~\cite{xiaomi2025mimo}. The target model is selected from either Qwen3-VL-32B (Q32B)~\cite{yang2025qwen3} or Qwen3-VL-235B-A22B (Q235B). All draft and target models are used in their thinking versions.

All draft and target model pairs use identical decoding settings. All evaluations are conducted on NVIDIA L40S GPUs. The target model is deployed on four L40S GPUs using AWQ INT4 quantization, while each draft model runs on two L40S GPUs. We use a fixed speculative lookahead window to 4 and set the acceptance threshold $\alpha$ to 0.7. For each benchmark, we report accuracy, acceptance rate, and speedup relative to target model autoregressive decoding.
  
  


The draft model is trained using SAPO on a mixture of multimodal reasoning datasets with step-level annotations, including Geo3K~\cite{lu2021inter}, OCR-VQA~\cite{mishra2019ocr}, and the ScienceQA~\cite{lu2022learn} dev split. Training is conducted on $8\times$NVIDIA H200 GPUs with BF16 precision. We adopt the AdamW optimizer with a peak learning rate of $1\times10^{-6}$, a batch size of 64, and train for 15 epochs, using a maximum sequence length of 8196 tokens. The target model is kept frozen throughout training, while the draft model is fully optimized.

To evaluate DREAM-R, we compare it against three baseline methods. The first, \textit{Standard SD}, applies conventional speculative decoding~\cite{leviathan2023fast} to LRMs to accelerate generation. We further compare against two recent speculative reasoning approaches, SpecReason~\cite{pan2025specreason} and LR~\cite{fu2025scaling}. Finally, to isolate the contribution of SAPO, we include a variant of DREAM-R without SAPO, termed \textit{DREAM-R-NS}, which incorporates CPN and FPSR but omits SAPO. Specifically, we adopt three evaluation metrics: model accuracy (Acc.), the acceptance rate of draft reasoning steps during verification (Acpt.), and execution latency speedup.


\subsection{Evaluation Results}
\label{sec:main-results}

Table~\ref{tab:vanilla} summarizes the autoregressive accuracy of all candidate models and serves as the baseline for evaluating speculative reasoning. The target models, Qwen3-VL-32B and Qwen3-VL-235B, exhibit strong baseline performance, achieving $76.00\%$ and $85.40\%$ on MathVerse, $83.40\%$ and $83.20\%$ on MMBench, and over $75.00\%$ on RealWorldQA. In contrast, draft models are substantially weaker, with Qwen3-VL-2B and Qwen3-VL-4B reaching only $53.94\%$ and $65.40\%$ on MathVerse, respectively. This performance gap highlights the challenge of effective speculative reasoning when draft models differ significantly from the target.

Table~\ref{tab:spec} demonstrates that our framework preserves accuracy close to the vanilla baseline while delivering substantial acceleration. Using Qwen3-VL-32B as the target and Qwen3-VL-2B as the draft, the supervised variant achieves $74.30\%$ on MathVerse and $80.32\%$ on MMBench, with speedups ranging from 1.8$\times$ to 2.2$\times$ across benchmarks. The RL-enhanced variant further improves acceptance, yielding speedups of up to 2.38$\times$ on MathVerse and 2.48$\times$ on MMBench and RealWorldQA, while maintaining high accuracy. In contrast, SpecReason underperforms in the VLM setting because its verification relies on discrete, score-based judgments that are brittle under multimodal uncertainty. Small draft models are particularly prone to hallucinations, and the resulting noisy scores can interfere with the target verifier. Moreover, discrete scoring often produces closely clustered or extreme values, making threshold-based accept/reject decisions highly unstable. These factors jointly lead to low acceptance rates and degraded final accuracy.
\begin{table}[t]
\centering
\small
\setlength{\tabcolsep}{3pt}
\renewcommand{\arraystretch}{1.2}

\caption{Model performance (\%) across four benchmarks.}
\label{tab:vanilla}
\resizebox{0.9\linewidth}{!}{
\begin{tabular}{c cccc}
\toprule
\textbf{Model} &
\textbf{MathVerse} &
\textbf{MMBench} &
\textbf{RealWorldQA} &
\textbf{MMMU} \\
\midrule
Q32B     & 76.00 & 83.40 & 75.55 & 77.85 \\
Q235B    & 85.40 & 83.20 & 75.17 & 71.59 \\
Q2B      & 53.94 & 62.90 & 63.29 & 59.27 \\
Q4B      & 65.40 & 71.00 & 68.98 & 59.60 \\
M7B-RL   & 69.80 & 78.50 & 58.80 & 66.80 \\
R4B      & 75.40 & 74.40 & 66.48 & 65.20 \\
\bottomrule
\end{tabular}
}
\end{table}

 \subsection{Ablation Studies}
\label{sec:ablation}
\subsubsection{Ablation on Reward Function Design}
\begin{figure}
    \centering
    \includegraphics[width=\columnwidth]{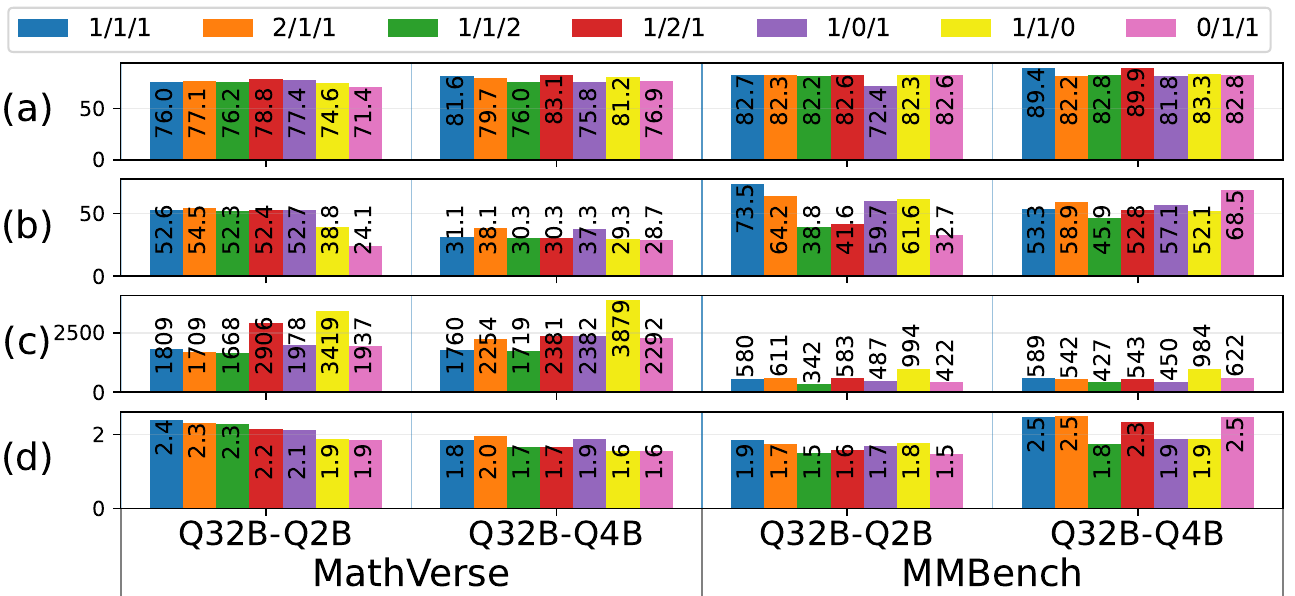}
    \caption{Impact of RL reward weighting on (a) output accuracy, (b) reasoning acceptance rate, (c) reasoning token length, and (d) speedup. Values separated by ``/'' indicate the weights $w_{1}$, $w_{2}$, $w_{3}$.}
    \label{fig:ab1}
\end{figure}

Based on Equation~\ref{eqn:reward-function}, we examine how different weighting schemes for draft alignment reward $R_{\text{draft}}$,  the outcome reward $R_{\text{outcome}}$, and length penalty reward term $R_{\text{length}}$ affect the behavior of the draft model. Unless stated otherwise, the baseline setting uses equal weights for all three components, and we vary the relative importance $w_{1},w_{2},w_{3}$ of each term to analyze its impact. According to Figure~\ref{fig:ab1}, the balanced weighting strategy with $w_{1}=1$, $w_{2}=1$, and $w_{3}=1$ delivers the best overall performance across benchmarks. Specifically, on MathVerse it achieves $75.98\%$ accuracy with a $52.61\%$ acceptance rate and a $2.38\times$ speedup, while on MMBench it attains $82.65\%$ accuracy with a $73.52\%$ acceptance rate and a $1.86\times$ speedup.

Increasing the weight $w_{2}$ on $R_{\text{draft}}$ while keeping other terms fixed leads to only marginal accuracy changes on MathVerse (77.12\%) and a modest increase in acceptance (54.54\%), but with reduced speedup ($2.31\times$). On MMBench, acceptance instead drops to $64.15\%$, suggesting that emphasizing verification alignment alone does not consistently improve verifier agreement in multimodal reasoning.

Adjusting the balance between $R_{\text{outcome}}$ and $R_{\text{length}}$ introduces more pronounced trade-offs. Emphasizing $R_{\text{outcome}}$ increases MathVerse accuracy to $78.80\%$, but substantially lengthens generation ($2906$ tokens) and reduces speedup ($2.16\times$). In contrast, strengthening $R_{\text{length}}$ shortens outputs on MathVerse ($1668$ tokens) but lowers acceptance to $52.26\%$ and causes a sharp acceptance drop on MMBench ($38.78\%$). These results indicate that while outcome-level supervision is beneficial, overly aggressive or imbalanced regularization can negatively affect verifier acceptance.

Employing a stronger draft model (Qwen3-VL-4B) further improves both accuracy and acceptance, resulting in more stable speculative decoding. Across all benchmarks, our approach consistently outperforms existing baselines: LR achieves moderate acceleration but often at the cost of accuracy, while SpecReason suffers from low acceptance when the draft is substantially weaker. In contrast, DREAM-R-NS and DREAM-R simultaneously maintains high accuracy and high acceptance, achieving the most favorable balance between efficiency and correctness.

Scaling to the larger Qwen3-VL-235B target shows that DREAM-R-NS and DREAM-R generalize well across model sizes. When using Qwen3-VL-2B as the draft, DREAM-R-NS achieves speedups ranging from 1.6$\times$ to 2.3$\times$ while maintaining accuracy close to the vanilla Qwen3-VL-235B baseline, which spans 71.59\% to 85.40\%. The RL-enhanced variant further improves acceptance by approximately $15\%$--$25\%$, leading to additional acceleration across all benchmarks. With Qwen3-VL-4B as the draft, accuracy closely matches the vanilla target performance on MMBench and RealWorldQA, while speculative efficiency remains robust, delivering speedups of 1.5$\times$ to 1.8$\times$.

Overall, the results demonstrate that our speculative decoding framework consistently achieves the best trade-off among accuracy retention, acceptance rate, and decoding efficiency. Both the supervised and RL-enhanced variants preserve accuracy close to the vanilla baseline in Table~1 while delivering substantial and stable speedups across four multimodal reasoning benchmarks.



\begin{table*}[h]
\centering
\small
\setlength{\tabcolsep}{2.2pt}
\renewcommand{\arraystretch}{1.02}

\caption{Performance evaluations across different models and datasets.}
\label{tab:spec}
\resizebox{\linewidth}{!}{
\begin{tabular}{
c c c
ccc ccc ccc ccc
}
\toprule
\multirow{2}{*}{\textbf{Target}} &
\multirow{2}{*}{\textbf{Draft}} &
\multirow{2}{*}{\textbf{Method}} &
\multicolumn{3}{c}{\textbf{MathVerse}} &
\multicolumn{3}{c}{\textbf{MMBench}} &
\multicolumn{3}{c}{\textbf{RealWorldQA}} &
\multicolumn{3}{c}{\textbf{MMMU}} \\
& & &
Acc.\% & Acpt.\% & Speedup &
Acc.\% & Acpt.\% & Speedup &
Acc.\% & Acpt.\% & Speedup &
Acc.\% & Acpt.\% & Speedup \\
\midrule

\multirow{20}{*}{Q32B}

& \multirow{5}{*}{Q2B}
& Standard SD      & 75.99 & --    & $1.09\,\times$  & 79.11 & --    & $1.15\,\times$  & 69.72 & --    & $1.12\,\times$  & 76.15 & --    & $1.15\,\times$ \\
& & SpecReason~\cite{pan2025specreason}   & 44.57 & 14.60 & $1.26\,\times$  & 62.49 & 24.26 & $1.43\,\times$  & 39.80 & 21.35 & $1.41\,\times$  & 46.88 & 16.04 & $1.21\,\times$ \\
& & LR~\cite{fu2025scaling}    & 72.36 & 35.21 & $1.53\,\times$  & 74.52 & 71.75 & $1.73\,\times$  & 72.97 & 66.85 & $2.13\,\times$  & 75.57 & 41.63 & $1.72\,\times$ \\
& & \cellcolor{LightLav}DREAM-R-NS
                   & \cellcolor{LightLav}74.30 & \cellcolor{LightLav}35.70 & \cellcolor{LightLav}{$1.86\,\times$}
                   & \cellcolor{LightLav}80.32 & \cellcolor{LightLav}69.99 & \cellcolor{LightLav}{$1.77\,\times$}
                   & \cellcolor{LightLav}75.55 & \cellcolor{LightLav}60.06 & \cellcolor{LightLav}{$2.19\,\times$}
                   & \cellcolor{LightLav}77.85 & \cellcolor{LightLav}42.46 & \cellcolor{LightLav}{$2.03\,\times$} \\
& & \cellcolor{LightLav}DREAM-R
                   & \cellcolor{LightLav}75.98 & \cellcolor{LightLav}52.61 & \cellcolor{LightLav}{$2.38\,\times$}
                   & \cellcolor{LightLav}92.65 & \cellcolor{LightLav}63.52 & \cellcolor{LightLav}{$1.86\,\times$}
                   & \cellcolor{LightLav}76.10 & \cellcolor{LightLav}67.98 & \cellcolor{LightLav}{$2.48\,\times$}
                   & \cellcolor{LightLav}85.79 & \cellcolor{LightLav}56.44 & \cellcolor{LightLav}{$2.19\,\times$} \\

\cline{2-15}

& \multirow{5}{*}{Q4B}
& Standard SD      & 71.31 & --    & $1.04\,\times$ & 83.90 & --    & $1.11\,\times$ & 73.48 & --    & $1.08\,\times$ & 77.18 & --    & $1.10\,\times$ \\
& & SpecReason   & 40.22 & 20.48 & $0.94\,\times$ & 73.58 & 32.13 & $1.28\,\times$ & 69.86 & 48.51 & $1.23\,\times$ & 53.73 & 32.86 & $1.24\,\times$ \\
& & LR    & 72.08 & 40.28 & $1.55\,\times$ & 76.41 & 48.92 & $1.58\,\times$ & 72.12 & 49.29 & $1.66\,\times$ & 58.96 & 34.35 & $1.66\,\times$ \\
& & \cellcolor{LightLav}DREAM-R-NS
                   & \cellcolor{LightLav}74.80 & \cellcolor{LightLav}37.33 & \cellcolor{LightLav}{$1.64\,\times$}
                   & \cellcolor{LightLav}79.42 & \cellcolor{LightLav}49.15 & \cellcolor{LightLav}{$2.37\,\times$}
                   & \cellcolor{LightLav}81.38 & \cellcolor{LightLav}47.43 & \cellcolor{LightLav}{$1.71\,\times$}
                   & \cellcolor{LightLav}66.00 & \cellcolor{LightLav}34.23 & \cellcolor{LightLav}{$1.84\,\times$} \\
& & \cellcolor{LightLav}DREAM-R
                   & \cellcolor{LightLav}81.60 & \cellcolor{LightLav}31.13 & \cellcolor{LightLav}{$1.84\,\times$}
                   & \cellcolor{LightLav}89.44 & \cellcolor{LightLav}53.28 & \cellcolor{LightLav}{$2.48\,\times$}
                   & \cellcolor{LightLav}81.39 & \cellcolor{LightLav}52.36 & \cellcolor{LightLav}{$1.80\,\times$}
                   & \cellcolor{LightLav}77.37 & \cellcolor{LightLav}42.37 & \cellcolor{LightLav}{$1.91\,\times$} \\

\cline{2-15}

& \multirow{5}{*}{M7B-RL}
& Standard SD      & 73.20 & --    & $1.02\,\times$ & 86.90 & --    & $1.05\,\times$ & 71.10 & --    & $1.03\,\times$ & 74.00 & --    & $1.04\,\times$ \\
& & SpecReason   & 48.30 & 18.20 & $0.86\,\times$ & 45.20 & 15.10 & $0.80\,\times$ & 51.60 & 23.40 & $0.83\,\times$ & 50.10 & 27.90 & $0.84\,\times$ \\
& & LR    & 72.10 & 32.50 & $1.40\,\times$ & 84.30 & 35.60 & $1.32\,\times$ & 70.50 & 39.70 & $1.38\,\times$ & 69.20 & 37.30 & $1.36\,\times$ \\
& & \cellcolor{LightLav}DREAM-R-NS
                   & \cellcolor{LightLav}76.50 & \cellcolor{LightLav}34.10 & \cellcolor{LightLav}{$1.55\,\times$}
                   & \cellcolor{LightLav}88.10 & \cellcolor{LightLav}60.20 & \cellcolor{LightLav}{$1.48\,\times$}
                   & \cellcolor{LightLav}73.00 & \cellcolor{LightLav}41.50 & \cellcolor{LightLav}{$1.52\,\times$}
                   & \cellcolor{LightLav}71.10 & \cellcolor{LightLav}39.80 & \cellcolor{LightLav}{$1.49\,\times$} \\
& & \cellcolor{LightLav}DREAM-R
                   & \cellcolor{LightLav}79.80 & \cellcolor{LightLav}36.70 & \cellcolor{LightLav}{$1.70\,\times$}
                   & \cellcolor{LightLav}90.40 & \cellcolor{LightLav}63.80 & \cellcolor{LightLav}{$1.62\,\times$}
                   & \cellcolor{LightLav}75.20 & \cellcolor{LightLav}44.00 & \cellcolor{LightLav}{$1.66\,\times$}
                   & \cellcolor{LightLav}73.50 & \cellcolor{LightLav}42.10 & \cellcolor{LightLav}{$1.61\,\times$} \\

\cline{2-15}

& \multirow{5}{*}{R4B}
& Standard SD      & 72.10 & --    & $1.07\,\times$ & 82.10 & --    & $1.09\,\times$ & 72.10 & --    & $1.09\,\times$ & 68.15 & --    & $1.08\,\times$ \\
& & SpecReason   & 46.35 & 21.80 & $1.27\,\times$ & 71.20 & 29.30 & $1.24\,\times$ & 62.10 & 38.70 & $1.26\,\times$ & 54.30 & 40.50 & $1.22\,\times$ \\
& & LR    & 71.25 & 43.90 & $1.67\,\times$ & 77.35 & 47.80 & $1.78\,\times$ & 70.95 & 53.80 & $1.82\,\times$ & 67.10 & 46.90 & $1.80\,\times$ \\
& & \cellcolor{LightLav}DREAM-R-NS
                   & \cellcolor{LightLav}74.10 & \cellcolor{LightLav}45.60 & \cellcolor{LightLav}{$1.83\,\times$}
                   & \cellcolor{LightLav}80.15 & \cellcolor{LightLav}52.60 & \cellcolor{LightLav}{$2.18\,\times$}
                   & \cellcolor{LightLav}72.80 & \cellcolor{LightLav}50.90 & \cellcolor{LightLav}{$1.93\,\times$}
                   & \cellcolor{LightLav}69.05 & \cellcolor{LightLav}49.20 & \cellcolor{LightLav}{$1.88\,\times$} \\
& & \cellcolor{LightLav}DREAM-R
                   & \cellcolor{LightLav}78.45 & \cellcolor{LightLav}49.70 & \cellcolor{LightLav}{$1.96\,\times$}
                   & \cellcolor{LightLav}83.60 & \cellcolor{LightLav}57.10 & \cellcolor{LightLav}{$2.31\,\times$}
                   & \cellcolor{LightLav}75.25 & \cellcolor{LightLav}56.10 & \cellcolor{LightLav}{$2.03\,\times$}
                   & \cellcolor{LightLav}71.40 & \cellcolor{LightLav}53.60 & \cellcolor{LightLav}{$1.98\,\times$} \\

\midrule

\multirow{20}{*}{Q235B}

& \multirow{5}{*}{Q2B}
& Standard SD      & 82.22 & --    & $1.11\,\times$ & 83.20 & --    & $1.06\,\times$ & 75.17 & --    & $1.16\,\times$ & 71.59 & --    & $1.07\,\times$ \\
& & SpecReason   & 55.10 & 41.55 & $1.47\,\times$ & 73.29 & 68.47 & $1.13\,\times$ & 67.87 & 68.75 & $1.30\,\times$ & 56.71 & 35.65 & $1.07\,\times$ \\
& & LR    & 66.47 & 48.53 & $1.71\,\times$ & 80.50 & 66.42 & $1.93\,\times$ & 74.51 & 52.12 & $1.61\,\times$ & 71.11 & 36.66 & $1.36\,\times$ \\
& & \cellcolor{LightLav}DREAM-R-NS
                   & \cellcolor{LightLav}72.70 & \cellcolor{LightLav}49.80 & \cellcolor{LightLav}{$2.22\,\times$}
                   & \cellcolor{LightLav}81.73 & \cellcolor{LightLav}66.80 & \cellcolor{LightLav}{$2.28\,\times$}
                   & \cellcolor{LightLav}75.17 & \cellcolor{LightLav}57.29 & \cellcolor{LightLav}{$2.13\,\times$}
                   & \cellcolor{LightLav}71.59 & \cellcolor{LightLav}37.20 & \cellcolor{LightLav}{$1.52\,\times$} \\
& & \cellcolor{LightLav}DREAM-R
                   & \cellcolor{LightLav}80.00 & \cellcolor{LightLav}51.82 & \cellcolor{LightLav}{$2.28\,\times$}
                   & \cellcolor{LightLav}88.45 & \cellcolor{LightLav}71.45 & \cellcolor{LightLav}{$2.31\,\times$}
                   & \cellcolor{LightLav}77.89 & \cellcolor{LightLav}68.05 & \cellcolor{LightLav}{$2.26\,\times$}
                   & \cellcolor{LightLav}78.33 & \cellcolor{LightLav}39.21 & \cellcolor{LightLav}{$1.60\,\times$} \\

\cline{2-15}

& \multirow{5}{*}{Q4B}
& Standard SD      & 81.24 & --    & $1.11\,\times$ & 81.31 & --    & $1.09\,\times$ & 75.36 & --    & $1.10\,\times$ & 69.83 & --    & $1.11\,\times$ \\
& & SpecReason   & 47.35 & 29.91 & $1.23\,\times$ & 69.74 & 39.18 & $1.22\,\times$ & 55.43 & 36.88 & $1.23\,\times$ & 59.72 & 51.86 & $1.22\,\times$ \\
& & LR    & 70.70 & 57.65 & $1.68\,\times$ & 79.10 & 37.34 & $1.45\,\times$ & 69.88 & 48.22 & $1.48\,\times$ & 68.20 & 53.70 & $1.48\,\times$ \\
& & \cellcolor{LightLav}DREAM-R-NS
                   & \cellcolor{LightLav}80.40 & \cellcolor{LightLav}56.94 & \cellcolor{LightLav}{$1.77\,\times$}
                   & \cellcolor{LightLav}81.32 & \cellcolor{LightLav}39.01 & \cellcolor{LightLav}{$1.65\,\times$}
                   & \cellcolor{LightLav}68.52 & \cellcolor{LightLav}44.23 & \cellcolor{LightLav}{$1.51\,\times$}
                   & \cellcolor{LightLav}71.69 & \cellcolor{LightLav}53.29 & \cellcolor{LightLav}{$1.55\,\times$} \\
& & \cellcolor{LightLav}DREAM-R
                   & \cellcolor{LightLav}84.00 & \cellcolor{LightLav}57.96 & \cellcolor{LightLav}{$1.79\,\times$}
                   & \cellcolor{LightLav}82.20 & \cellcolor{LightLav}37.96 & \cellcolor{LightLav}{$1.73\,\times$}
                   & \cellcolor{LightLav}72.53 & \cellcolor{LightLav}47.22 & \cellcolor{LightLav}{$1.53\,\times$}
                   & \cellcolor{LightLav}72.53 & \cellcolor{LightLav}57.82 & \cellcolor{LightLav}{$1.58\,\times$} \\

\cline{2-15}

& \multirow{5}{*}{M7B-RL}
& Standard SD      & 81.20 & --    & $1.08\,\times$ & 82.70 & --    & $1.04\,\times$ & 75.10 & --    & $1.07\,\times$ & 71.50 & --    & $1.07\,\times$ \\
& & SpecReason   & 64.40 & 36.52 & $0.80\,\times$ & 17.10 & 14.82 & $0.71\,\times$ & 11.30 & 24.79 & $1.09\,\times$ & 48.60 & 23.77 & $0.63\,\times$ \\
& & LR    & 72.10 & 37.91 & $1.46\,\times$ & 74.70 & 35.46 & $1.46\,\times$ & 73.30 & 25.24 & $1.40\,\times$ & 79.80 & 23.03 & $1.45\,\times$ \\
& & \cellcolor{LightLav}DREAM-R-NS
                   & \cellcolor{LightLav}78.00 & \cellcolor{LightLav}36.80 & \cellcolor{LightLav}{$1.62\,\times$}
                   & \cellcolor{LightLav}79.60 & \cellcolor{LightLav}35.45 & \cellcolor{LightLav}{$1.62\,\times$}
                   & \cellcolor{LightLav}76.10 & \cellcolor{LightLav}25.56 & \cellcolor{LightLav}{$1.56\,\times$}
                   & \cellcolor{LightLav}81.00 & \cellcolor{LightLav}23.69 & \cellcolor{LightLav}{$1.65\,\times$} \\
& & \cellcolor{LightLav}DREAM-R
                   & \cellcolor{LightLav}82.10 & \cellcolor{LightLav}40.20 & \cellcolor{LightLav}{$1.78\,\times$}
                   & \cellcolor{LightLav}82.30 & \cellcolor{LightLav}47.00 & \cellcolor{LightLav}{$1.78\,\times$}
                   & \cellcolor{LightLav}78.40 & \cellcolor{LightLav}28.50 & \cellcolor{LightLav}{$1.72\,\times$}
                   & \cellcolor{LightLav}83.10 & \cellcolor{LightLav}25.00 & \cellcolor{LightLav}{$1.75\,\times$} \\

\cline{2-15}

& \multirow{5}{*}{R4B}
& Standard SD      & 80.10 & --    & $1.10\,\times$ & 80.95 & --    & $1.08\,\times$ & 74.20 & --    & $1.09\,\times$ & 68.05 & --    & $1.10\,\times$ \\
& & SpecReason   & 49.20 & 27.40 & $1.21\,\times$ & 67.10 & 37.50 & $1.20\,\times$ & 57.10 & 35.60 & $1.21\,\times$ & 60.10 & 49.80 & $1.21\,\times$ \\
& & LR    & 69.85 & 54.20 & $1.65\,\times$ & 78.25 & 35.80 & $1.43\,\times$ & 69.05 & 46.90 & $1.47\,\times$ & 67.30 & 51.90 & $1.46\,\times$ \\
& & \cellcolor{LightLav}DREAM-R-NS
                   & \cellcolor{LightLav}79.10 & \cellcolor{LightLav}53.70 & \cellcolor{LightLav}{$1.74\,\times$}
                   & \cellcolor{LightLav}79.85 & \cellcolor{LightLav}37.90 & \cellcolor{LightLav}{$1.62\,\times$}
                   & \cellcolor{LightLav}67.95 & \cellcolor{LightLav}42.70 & \cellcolor{LightLav}{$1.49\,\times$}
                   & \cellcolor{LightLav}70.25 & \cellcolor{LightLav}50.70 & \cellcolor{LightLav}{$1.54\,\times$} \\
& & \cellcolor{LightLav}DREAM-R
                   & \cellcolor{LightLav}82.35 & \cellcolor{LightLav}55.90 & \cellcolor{LightLav}{$1.78\,\times$}
                   & \cellcolor{LightLav}81.95 & \cellcolor{LightLav}36.90 & \cellcolor{LightLav}{$1.71\,\times$}
                   & \cellcolor{LightLav}71.80 & \cellcolor{LightLav}45.90 & \cellcolor{LightLav}{$1.52\,\times$}
                   & \cellcolor{LightLav}71.95 & \cellcolor{LightLav}54.10 & \cellcolor{LightLav}{$1.57\,\times$} \\

\bottomrule
\end{tabular}
 }
\end{table*}

\subsubsection{Ablation on FPSR}
\label{sec:ablation_fpsr}
\begin{figure}
    \centering
    \includegraphics[width=\columnwidth]{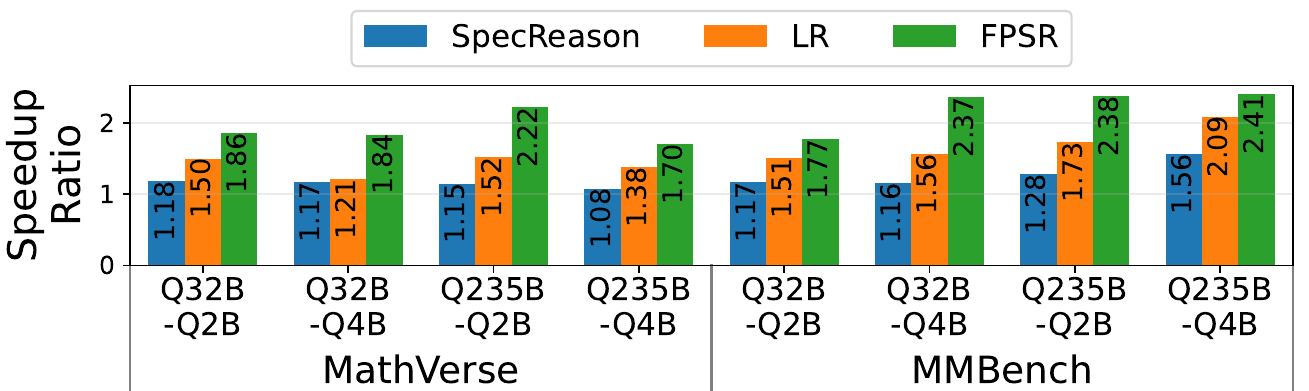}
    \caption{Impact of different scheduling methods.}
    \label{fig:ab2}
\end{figure}
\begin{figure}
    \centering
    \includegraphics[width=\columnwidth]{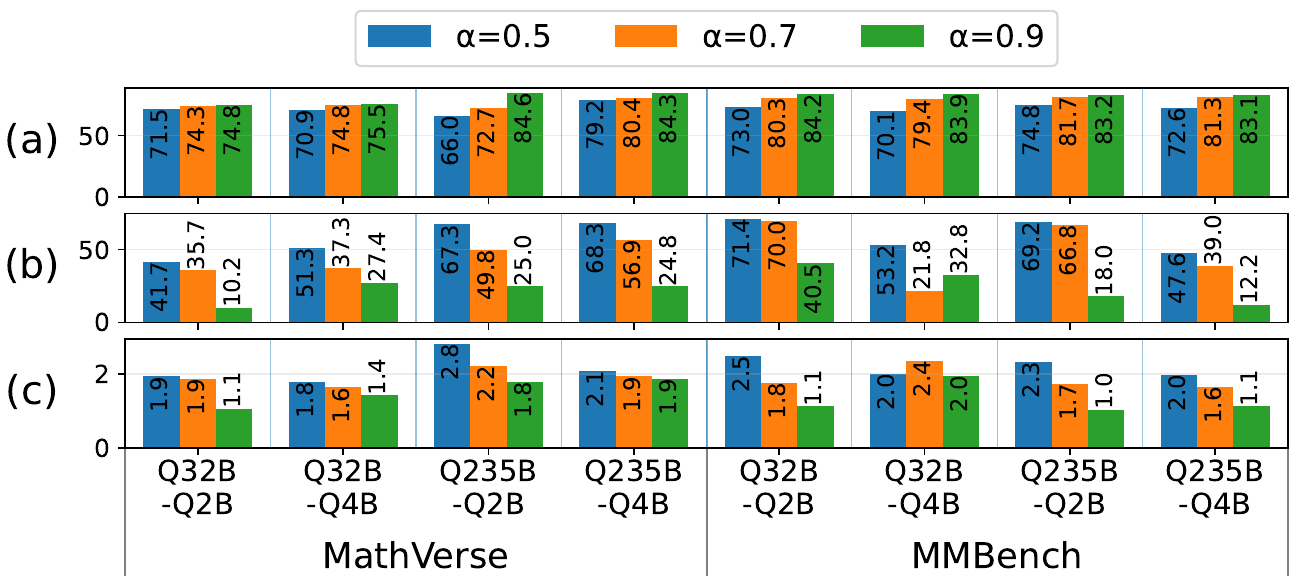}
    \caption{Impact of $\alpha$ on (a) accuracy (b) reasoning acceptance rate and (c) speedup. }
    \label{fig:ab3}
\end{figure}


We analyze the impact of FPSR by comparing it with two baselines: SpecReason~\cite{pan2025specreason}, LR~\cite{fu2025scaling}, and FPSR. SpecReason does not incorporate explicit scheduling mechanisms and instead directly follows the baseline scheduling strategy illustrated in Figure~\ref{fig:fpsr}(a). In contrast, LR allows the draft and target models to generate in parallel, overlapping their computation to improve hardware utilization and reduce end-to-end latency. However, verification is deferred until generation finishes, delaying feedback and preventing early rejection of incorrect draft steps. As shown in Figure~\ref{fig:ab2}, for the Qwen3-VL-2B to Qwen3-VL-32B setting, speedup increases from 1.18$\times$ and 1.17$\times$ with SpecReason~\cite{pan2025specreason} on MathVerse and MMBench to 1.50$\times$ and 1.51$\times$ with FPSR, and further to 1.86$\times$ and 1.77$\times$ under full parallelism. With a stronger draft model (Qwen3-VL-4B), FPSR achieves speedups of 1.84$\times$ and 2.37$\times$ on the two benchmarks.
When the target is Qwen3-VL-235B, the benefit becomes even more significant. For example, with a 2B draft, speedup increases from 1.15$\times$ and 1.28$\times$ to 1.52$\times$ and 1.73$\times$ under LR~\cite{fu2025scaling}, and to 2.22$\times$ and 2.38$\times$ under FPSR. 


\subsubsection{Ablation on Acceptance Threshold for CPN}

We analyze how the threshold $\alpha$ in CPN (Section~\ref{sec:tbvm}) affects the performance of DREAM-R. By comparing $\phi$ with $\alpha$, this threshold controls the acceptance ratio of drafted reasoning steps. A lower $\alpha$ makes the verifier more permissive, leading to higher acceptance rates and the largest speedups across model pairs, but at the cost of reduced accuracy due to accepting lower-quality reasoning. In contrast, a higher $\alpha$ makes the verifier more conservative, improving accuracy while increasing latency. As shown in Figure~\ref{fig:ab3}, setting $\alpha$ to 0.7 achieves the best overall trade-off between accuracy and speedup across different models and datasets. This indicates that the acceptance threshold serves as an effective control knob for tuning system behavior: lower thresholds prioritize efficiency, higher thresholds prioritize accuracy, and intermediate thresholds strike a strong balance.

\section{Conclusion}
\label{sec:conclusion}
We propose \textit{DREAM-R}, a multimodal speculative reasoning framework that accelerates reasoning-heavy decoding while preserving target-model accuracy. Experiments on four multimodal reasoning benchmarks show up to $2.48\times$ speedup with accuracy comparable to vanilla target decoding, outperforming prior speculative reasoning baselines.

\section*{Impact Statement}
\textit{DREAM-R} aims to improve the efficiency and scalability of multimodal reasoning systems through speculative reasoning and reinforcement learning based alignment techniques. By reducing inference latency and computational cost, the proposed approach may help make advanced AI systems more accessible and energy-efficient.

At the same time, the underlying foundation models may still inherit limitations such as hallucinations, biased outputs, or incorrect reasoning. As with other large-scale AI systems, the techniques presented in this work could potentially be integrated into high-throughput automated decision-making or content generation pipelines. We therefore encourage responsible deployment and appropriate human oversight when applying such systems in real-world settings.

\clearpage
\nocite{langley00}

\bibliography{custom}

@inproceedings{lu2022learn,
    title={Learn to Explain: Multimodal Reasoning via Thought Chains for Science Question Answering},
    author={Lu, Pan and Mishra, Swaroop and Xia, Tony and Qiu, Liang and Chang, Kai-Wei and Zhu, Song-Chun and Tafjord, Oyvind and Clark, Peter and Ashwin Kalyan},
    booktitle={The 36th Conference on Neural Information Processing Systems (NeurIPS)},
    year={2022}
}

@article{yang2025r,
  title={R-4b: Incentivizing general-purpose auto-thinking capability in mllms via bi-mode annealing and reinforce learning},
  author={Yang, Qi and Ni, Bolin and Xiang, Shiming and Hu, Han and Peng, Houwen and Jiang, Jie},
  journal={arXiv preprint arXiv:2508.21113},
  year={2025}
}

@article{shao2024deepseekmath,
  title={Deepseekmath: Pushing the limits of mathematical reasoning in open language models},
  author={Shao, Zhihong and Wang, Peiyi and Zhu, Qihao and Xu, Runxin and Song, Junxiao and Bi, Xiao and Zhang, Haowei and Zhang, Mingchuan and Li, YK and Wu, Yang and others},
  journal={arXiv preprint arXiv:2402.03300},
  year={2024}
}

@article{hu2025dream,
  title={DREAM: Drafting with Refined Target Features and Entropy-Adaptive Cross-Attention Fusion for Multimodal Speculative Decoding},
  author={Hu, Yunhai and Xia, Tianhua and Liu, Zining and Raman, Rahul and Liu, Xingyu and Bao, Bo and Sather, Eric and Thangarasa, Vithursan and Zhang, Sai Qian},
  journal={arXiv preprint arXiv:2505.19201},
  year={2025}
}

@article{pan2025specreason,
  title={Specreason: Fast and accurate inference-time compute via speculative reasoning},
  author={Pan, Rui and Dai, Yinwei and Zhang, Zhihao and Oliaro, Gabriele and Jia, Zhihao and Netravali, Ravi},
  journal={arXiv preprint arXiv:2504.07891},
  year={2025}
}

@article{fu2025scaling,
  title={Scaling Speculative Decoding with Lookahead Reasoning},
  author={Fu, Yichao and Ge, Rui and Shao, Zelei and Deng, Zhijie and Zhang, Hao},
  journal={arXiv preprint arXiv:2506.19830},
  year={2025}
}

@inproceedings{leviathan2023fast,
  title={Fast inference from transformers via speculative decoding},
  author={Leviathan, Yaniv and Kalman, Matan and Matias, Yossi},
  booktitle={International Conference on Machine Learning},
  pages={19274--19286},
  year={2023},
  organization={PMLR}
}

@article{ouyang2022training,
  title={Training language models to follow instructions with human feedback},
  author={Ouyang, Long and Wu, Jeffrey and Jiang, Xu and Almeida, Diogo and Wainwright, Carroll and Mishkin, Pamela and Zhang, Chong and Agarwal, Sandhini and Slama, Katarina and Ray, Alex and others},
  journal={Advances in neural information processing systems},
  volume={35},
  pages={27730--27744},
  year={2022}
}

@article{brown2020language,
  title={Language models are few-shot learners},
  author={Brown, Tom and Mann, Benjamin and Ryder, Nick and Subbiah, Melanie and Kaplan, Jared D and Dhariwal, Prafulla and Neelakantan, Arvind and Shyam, Pranav and Sastry, Girish and Askell, Amanda and others},
  journal={Advances in neural information processing systems},
  volume={33},
  pages={1877--1901},
  year={2020}
}

@article{achiam2023gpt,
  title={Gpt-4 technical report},
  author={Achiam, Josh and Adler, Steven and Agarwal, Sandhini and Ahmad, Lama and Akkaya, Ilge and Aleman, Florencia Leoni and Almeida, Diogo and Altenschmidt, Janko and Altman, Sam and Anadkat, Shyamal and others},
  journal={arXiv preprint arXiv:2303.08774},
  year={2023}
}

@inproceedings{zhang2024mathverse,
  title={Mathverse: Does your multi-modal llm truly see the diagrams in visual math problems?},
  author={Zhang, Renrui and Jiang, Dongzhi and Zhang, Yichi and Lin, Haokun and Guo, Ziyu and Qiu, Pengshuo and Zhou, Aojun and Lu, Pan and Chang, Kai-Wei and Qiao, Yu and others},
  booktitle={European Conference on Computer Vision},
  pages={169--186},
  year={2024},
  organization={Springer}
}

@inproceedings{mishra2019ocr,
  title={Ocr-vqa: Visual question answering by reading text in images},
  author={Mishra, Anand and Shekhar, Shashank and Singh, Ajeet Kumar and Chakraborty, Anirban},
  booktitle={2019 international conference on document analysis and recognition (ICDAR)},
  pages={947--952},
  year={2019},
  organization={IEEE}
}

@article{lu2021inter,
  title={Inter-gps: Interpretable geometry problem solving with formal language and symbolic reasoning},
  author={Lu, Pan and Gong, Ran and Jiang, Shibiao and Qiu, Liang and Huang, Siyuan and Liang, Xiaodan and Zhu, Song-Chun},
  journal={arXiv preprint arXiv:2105.04165},
  year={2021}
}

@article{xiaomi2025mimo,
  title={MiMo: Unlocking the Reasoning Potential of Language Model--From Pretraining to Posttraining},
  author={Xiaomi, LLM and Xia, Bingquan and Shen, Bowen and Zhu, Dawei and Zhang, Di and Wang, Gang and Zhang, Hailin and Liu, Huaqiu and Xiao, Jiebao and Dong, Jinhao and others},
  journal={arXiv preprint arXiv:2505.07608},
  year={2025}
}

@article{yang2025qwen3,
  title={Qwen3 technical report},
  author={Yang, An and Li, Anfeng and Yang, Baosong and Zhang, Beichen and Hui, Binyuan and Zheng, Bo and Yu, Bowen and Gao, Chang and Huang, Chengen and Lv, Chenxu and others},
  journal={arXiv preprint arXiv:2505.09388},
  year={2025}
}

@inproceedings{liu2024mmbench,
  title={Mmbench: Is your multi-modal model an all-around player?},
  author={Liu, Yuan and Duan, Haodong and Zhang, Yuanhan and Li, Bo and Zhang, Songyang and Zhao, Wangbo and Yuan, Yike and Wang, Jiaqi and He, Conghui and Liu, Ziwei and others},
  booktitle={European conference on computer vision},
  pages={216--233},
  year={2024},
  organization={Springer}
}

@inproceedings{yue2024mmmu,
  title={Mmmu: A massive multi-discipline multimodal understanding and reasoning benchmark for expert agi},
  author={Yue, Xiang and Ni, Yuansheng and Zhang, Kai and Zheng, Tianyu and Liu, Ruoqi and Zhang, Ge and Stevens, Samuel and Jiang, Dongfu and Ren, Weiming and Sun, Yuxuan and others},
  booktitle={Proceedings of the IEEE/CVF Conference on Computer Vision and Pattern Recognition},
  pages={9556--9567},
  year={2024}
}

@misc{realworldqa2024,
  title        = {RealWorldQA Dataset},
  author       = {{xAI}},
  year         = {2024},
  howpublished = {\url{https://huggingface.co/datasets/xai-org/RealworldQA}},
  note         = {Version 1, CC BY-ND 4.0 License}
}

@article{liu2024deepseek,
  title={Deepseek-v3 technical report},
  author={Liu, Aixin and Feng, Bei and Xue, Bing and Wang, Bingxuan and Wu, Bochao and Lu, Chengda and Zhao, Chenggang and Deng, Chengqi and Zhang, Chenyu and Ruan, Chong and others},
  journal={arXiv preprint arXiv:2412.19437},
  year={2024}
}

@article{grattafiori2024llama,
  title={The llama 3 herd of models},
  author={Grattafiori, Aaron and Dubey, Abhimanyu and Jauhri, Abhinav and Pandey, Abhinav and Kadian, Abhishek and Al-Dahle, Ahmad and Letman, Aiesha and Mathur, Akhil and Schelten, Alan and Vaughan, Alex and others},
  journal={arXiv preprint arXiv:2407.21783},
  year={2024}
}

@inproceedings{li2025fast,
  title={Fast and High-Quality Auto-Regressive Speech Synthesis via Speculative Decoding},
  author={Li, Bohan and Wang, Hankun and Zhang, Situo and Guo, Yiwei and Yu, Kai},
  booktitle={ICASSP 2025-2025 IEEE International Conference on Acoustics, Speech and Signal Processing (ICASSP)},
  pages={1--5},
  year={2025},
  organization={IEEE}
}

@inproceedings{gagrani2024speculative,
  title={On speculative decoding for multimodal large language models},
  author={Gagrani, Mukul and Goel, Raghavv and Jeon, Wonseok and Park, Junyoung and Lee, Mingu and Lott, Christopher},
  booktitle={Proceedings of the IEEE/CVF Conference on Computer Vision and Pattern Recognition},
  pages={8285--8289},
  year={2024}
}

@inproceedings{
lee2025inbatch,
title={In-batch Ensemble Drafting: Robust Speculative Decoding for {LVLM}s},
author={Minjae Lee and Wonjun Kang and Byeongkeun Ahn and Christian Classen and Minghao Yan and Hyung Il Koo and Kangwook Lee},
booktitle={First Workshop on Scalable Optimization for Efficient and Adaptive Foundation Models},
year={2025},
url={https://openreview.net/forum?id=ffDhpmwqdu}
}

@article{kang2025vispec,
  title={Vispec: Accelerating vision-language models with vision-aware speculative decoding},
  author={Kang, Jialiang and Shu, Han and Li, Wenshuo and Zhai, Yingjie and Chen, Xinghao},
  journal={arXiv preprint arXiv:2509.15235},
  year={2025}
}

@article{ganesan2025massv,
  title={Massv: Multimodal adaptation and self-data distillation for speculative decoding of vision-language models},
  author={Ganesan, Mugilan and Segal, Shane and Aggarwal, Ankur and Sinnadurai, Nish and Lie, Sean and Thangarasa, Vithursan},
  journal={arXiv preprint arXiv:2505.10526},
  year={2025}
}

@article{li2024eagle2,
  title={Eagle-2: Faster inference of language models with dynamic draft trees},
  author={Li, Yuhui and Wei, Fangyun and Zhang, Chao and Zhang, Hongyang},
  journal={arXiv preprint arXiv:2406.16858},
  year={2024}
}

@article{li2024eagle,
  title={Eagle: Speculative sampling requires rethinking feature uncertainty},
  author={Li, Yuhui and Wei, Fangyun and Zhang, Chao and Zhang, Hongyang},
  journal={arXiv preprint arXiv:2401.15077},
  year={2024}
}

@article{hong2025glm,
  title={GLM-4.1 V-Thinking: Towards Versatile Multimodal Reasoning with Scalable Reinforcement Learning},
  author={Hong, Wenyi and Yu, Wenmeng and Gu, Xiaotao and Wang, Guo and Gan, Guobing and Tang, Haomiao and Cheng, Jiale and Qi, Ji and Ji, Junhui and Pan, Lihang and others},
  journal={arXiv preprint arXiv:2507.01006},
  year={2025}
}

@article{wang2025internvl3,
  title={Internvl3. 5: Advancing open-source multimodal models in versatility, reasoning, and efficiency},
  author={Wang, Weiyun and Gao, Zhangwei and Gu, Lixin and Pu, Hengjun and Cui, Long and Wei, Xingguang and Liu, Zhaoyang and Jing, Linglin and Ye, Shenglong and Shao, Jie and others},
  journal={arXiv preprint arXiv:2508.18265},
  year={2025}
}

@article{bai2025qwen2,
  title={Qwen2. 5-vl technical report},
  author={Bai, Shuai and Chen, Keqin and Liu, Xuejing and Wang, Jialin and Ge, Wenbin and Song, Sibo and Dang, Kai and Wang, Peng and Wang, Shijie and Tang, Jun and others},
  journal={arXiv preprint arXiv:2502.13923},
  year={2025}
}

@article{cobbe2021training,
  title={Training verifiers to solve math word problems},
  author={Cobbe, Karl and Kosaraju, Vineet and Bavarian, Mohammad and Chen, Mark and Jun, Heewoo and Kaiser, Lukasz and Plappert, Matthias and Tworek, Jerry and Hilton, Jacob and Nakano, Reiichiro and others},
  journal={arXiv preprint arXiv:2110.14168},
  year={2021}
}

@article{ou2024lossless,
  title={Lossless Acceleration of Large Language Model via Adaptive N-gram Parallel Decoding},
  author={Ou, Jie and Chen, Yueming and Tian, Wenhong},
  journal={arXiv preprint arXiv:2404.08698},
  year={2024}
}

@article{chen2021evaluating,
  title={Evaluating large language models trained on code},
  author={Chen, Mark},
  journal={arXiv preprint arXiv:2107.03374},
  year={2021}
}

@article{stewart2024n,
  title={The N-Grammys: Accelerating Autoregressive Inference with Learning-Free Batched Speculation},
  author={Stewart, Lawrence and Trager, Matthew and Gonugondla, Sujan Kumar and Soatto, Stefano},
  journal={arXiv preprint arXiv:2411.03786},
  year={2024}
}

@article{wei2022chain,
  title={Chain-of-thought prompting elicits reasoning in large language models},
  author={Wei, Jason and Wang, Xuezhi and Schuurmans, Dale and Bosma, Maarten and Xia, Fei and Chi, Ed and Le, Quoc V and Zhou, Denny and others},
  journal={Advances in neural information processing systems},
  volume={35},
  pages={24824--24837},
  year={2022}
}

@article{gao2024falcon,
  title={Falcon: Faster and Parallel Inference of Large Language Models through Enhanced Semi-Autoregressive Drafting and Custom-Designed Decoding Tree},
  author={Gao, Xiangxiang and Xie, Weisheng and Xiang, Yiwei and Ji, Feng},
  journal={arXiv preprint arXiv:2412.12639},
  year={2024}
}

@article{elhoushi2024layer,
  title={Layer skip: Enabling early exit inference and self-speculative decoding},
  author={Elhoushi, Mostafa and Shrivastava, Akshat and Liskovich, Diana and Hosmer, Basil and Wasti, Bram and Lai, Liangzhen and Mahmoud, Anas and Acun, Bilge and Agarwal, Saurabh and Roman, Ahmed and others},
  journal={arXiv preprint arXiv:2404.16710},
  year={2024}
}

@article{chen2023cascade,
  title={Cascade speculative drafting for even faster llm inference},
  author={Chen, Ziyi and Yang, Xiaocong and Lin, Jiacheng and Sun, Chenkai and Chang, Kevin Chen-Chuan and Huang, Jie},
  journal={arXiv preprint arXiv:2312.11462},
  year={2023}
}

@article{mcdanel2025pipespec,
  title={PipeSpec: Breaking Stage Dependencies in Hierarchical LLM Decoding},
  author={McDanel, Bradley and Zhang, Sai Qian and Hu, Yunhai and Liu, Zining},
  journal={arXiv preprint arXiv:2505.01572},
  year={2025}
}

@article{xia2024swift,
  title={SWIFT: On-the-Fly Self-Speculative Decoding for LLM Inference Acceleration},
  author={Xia, Heming and Li, Yongqi and Zhang, Jun and Du, Cunxiao and Li, Wenjie},
  journal={arXiv preprint arXiv:2410.06916},
  year={2024}
}

@article{yang2023inference,
  title={Inference with reference: Lossless acceleration of large language models},
  author={Yang, Nan and Ge, Tao and Wang, Liang and Jiao, Binxing and Jiang, Daxin and Yang, Linjun and Majumder, Rangan and Wei, Furu},
  journal={arXiv preprint arXiv:2304.04487},
  year={2023}
}

@article{he2023rest,
  title={Rest: Retrieval-based speculative decoding},
  author={He, Zhenyu and Zhong, Zexuan and Cai, Tianle and Lee, Jason D and He, Di},
  journal={arXiv preprint arXiv:2311.08252},
  year={2023}
}

@inproceedings{xia2023speculative,
  title={Speculative decoding: Exploiting speculative execution for accelerating seq2seq generation},
  author={Xia, Heming and Ge, Tao and Wang, Peiyi and Chen, Si-Qing and Wei, Furu and Sui, Zhifang},
  booktitle={Findings of the Association for Computational Linguistics: EMNLP 2023},
  pages={3909--3925},
  year={2023}
}

@article{hu2025speculative,
  title={Speculative Decoding and Beyond: An In-Depth Survey of Techniques},
  author={Hu, Yunhai and Liu, Zining and Dong, Zhenyuan and Peng, Tianfan and McDanel, Bradley and Zhang, Sai Qian},
  journal={arXiv preprint arXiv:2502.19732},
  year={2025}
}

@article{stern2018blockwise,
  title={Blockwise parallel decoding for deep autoregressive models},
  author={Stern, Mitchell and Shazeer, Noam and Uszkoreit, Jakob},
  journal={Advances in Neural Information Processing Systems},
  volume={31},
  year={2018}
}

@article{miao2023specinfer,
  title={SpecInfer: Accelerating Generative Large Language Model Serving with Tree-based Speculative Inference and Verification},
  author={Miao, Xupeng and Oliaro, Gabriele and Zhang, Zhihao and Cheng, Xinhao and Wang, Zeyu and Zhang, Zhengxin and Wong, Rae Ying Yee and Zhu, Alan and Yang, Lijie and Shi, Xiaoxiang and others},
  journal={arXiv preprint arXiv:2305.09781},
  year={2023}
}

@article{chen2024sequoia,
  title={Sequoia: Scalable, robust, and hardware-aware speculative decoding},
  author={Chen, Zhuoming and May, Avner and Svirschevski, Ruslan and Huang, Yuhsun and Ryabinin, Max and Jia, Zhihao and Chen, Beidi},
  journal={arXiv preprint arXiv:2402.12374},
  year={2024}
}

@article{li2025eagle3,
  title={EAGLE-3: Scaling up Inference Acceleration of Large Language Models via Training-Time Test},
  author={Li, Yuhui and Wei, Fangyun and Zhang, Chao and Zhang, Hongyang},
  journal={arXiv preprint arXiv:2503.01840},
  year={2025}
}

@article{liu2024kangaroo,
  title={Kangaroo: Lossless self-speculative decoding via double early exiting},
  author={Liu, Fangcheng and Tang, Yehui and Liu, Zhenhua and Ni, Yunsheng and Han, Kai and Wang, Yunhe},
  journal={arXiv preprint arXiv:2404.18911},
  year={2024}
}

@article{cai2024medusa,
  title={Medusa: Simple llm inference acceleration framework with multiple decoding heads},
  author={Cai, Tianle and Li, Yuhong and Geng, Zhengyang and Peng, Hongwu and Lee, Jason D and Chen, Deming and Dao, Tri},
  journal={arXiv preprint arXiv:2401.10774},
  year={2024}
}

@article{ankner2024hydra,
  title={Hydra: Sequentially-dependent draft heads for medusa decoding},
  author={Ankner, Zachary and Parthasarathy, Rishab and Nrusimha, Aniruddha and Rinard, Christopher and Ragan-Kelley, Jonathan and Brandon, William},
  journal={arXiv preprint arXiv:2402.05109},
  year={2024}
}

@article{sun2024triforce,
  title={Triforce: Lossless acceleration of long sequence generation with hierarchical speculative decoding},
  author={Sun, Hanshi and Chen, Zhuoming and Yang, Xinyu and Tian, Yuandong and Chen, Beidi},
  journal={arXiv preprint arXiv:2404.11912},
  year={2024}
}

@article{zhang2023draft,
  title={Draft \& verify: Lossless large language model acceleration via self-speculative decoding},
  author={Zhang, Jun and Wang, Jue and Li, Huan and Shou, Lidan and Chen, Ke and Chen, Gang and Mehrotra, Sharad},
  journal={arXiv preprint arXiv:2309.08168},
  year={2023}
}

@article{comanici2025gemini,
  title={Gemini 2.5: Pushing the frontier with advanced reasoning, multimodality, long context, and next generation agentic capabilities},
  author={Comanici, Gheorghe and Bieber, Eric and Schaekermann, Mike and Pasupat, Ice and Sachdeva, Noveen and Dhillon, Inderjit and Blistein, Marcel and Ram, Ori and Zhang, Dan and Rosen, Evan and others},
  journal={arXiv preprint arXiv:2507.06261},
  year={2025}
}

@article{guo2025deepseek,
  title={Deepseek-r1: Incentivizing reasoning capability in llms via reinforcement learning},
  author={Guo, Daya and Yang, Dejian and Zhang, Haowei and Song, Junxiao and Zhang, Ruoyu and Xu, Runxin and Zhu, Qihao and Ma, Shirong and Wang, Peiyi and Bi, Xiao and others},
  journal={arXiv preprint arXiv:2501.12948},
  year={2025}
}

@article{bai2022constitutional,
  title={Constitutional ai: Harmlessness from ai feedback},
  author={Bai, Yuntao and Kadavath, Saurav and Kundu, Sandipan and Askell, Amanda and Kernion, Jackson and Jones, Andy and Chen, Anna and Goldie, Anna and Mirhoseini, Azalia and McKinnon, Cameron and others},
  journal={arXiv preprint arXiv:2212.08073},
  year={2022}
}

@article{rafailov2023direct,
  title={Direct preference optimization: Your language model is secretly a reward model},
  author={Rafailov, Rafael and Sharma, Archit and Mitchell, Eric and Manning, Christopher D and Ermon, Stefano and Finn, Chelsea},
  journal={Advances in neural information processing systems},
  volume={36},
  pages={53728--53741},
  year={2023}
}

@article{ethayarajh2024kto,
  title={Kto: Model alignment as prospect theoretic optimization},
  author={Ethayarajh, Kawin and Xu, Winnie and Muennighoff, Niklas and Jurafsky, Dan and Kiela, Douwe},
  journal={arXiv preprint arXiv:2402.01306},
  year={2024}
}

@article{Ziegler2019FineTuningLM,
  title={Fine-Tuning Language Models from Human Preferences},
  author={Daniel M. Ziegler and Nisan Stiennon and Jeff Wu and Tom B. Brown and Alec Radford and Dario Amodei and Paul Christiano and Geoffrey Irving},
  journal={ArXiv},
  year={2019},
  volume={abs/1909.08593},
  url={https://api.semanticscholar.org/CorpusID:202660943}
}

@article{Stiennon2020LearningTS,
  title={Learning to summarize from human feedback},
  author={Nisan Stiennon and Long Ouyang and Jeff Wu and Daniel M. Ziegler and Ryan J. Lowe and Chelsea Voss and Alec Radford and Dario Amodei and Paul Christiano},
  journal={ArXiv},
  year={2020},
  volume={abs/2009.01325},
  url={https://api.semanticscholar.org/CorpusID:221665105}
}

@article{Sun2023AligningLM,
  title={Aligning Large Multimodal Models with Factually Augmented RLHF},
  author={Zhiqing Sun and Sheng Shen and Shengcao Cao and Haotian Liu and Chunyuan Li and Yikang Shen and Chuang Gan and Liangyan Gui and Yu-Xiong Wang and Yiming Yang and Kurt Keutzer and Trevor Darrell},
  journal={ArXiv},
  year={2023},
  volume={abs/2309.14525},
  url={https://api.semanticscholar.org/CorpusID:262824780}
}

@article{Gao2024OnDE,
  title={On Designing Effective RL Reward at Training Time for LLM Reasoning},
  author={Jiaxuan Gao and Shusheng Xu and Wenjie Ye and Weiling Liu and Chuyi He and Wei Fu and Zhiyu Mei and Guangju Wang and Yi Wu},
  journal={ArXiv},
  year={2024},
  volume={abs/2410.15115},
  url={https://api.semanticscholar.org/CorpusID:273502123}
}

@article{Zeng2025SimpleRLZooIA,
  title={SimpleRL-Zoo: Investigating and Taming Zero Reinforcement Learning for Open Base Models in the Wild},
  author={Weihao Zeng and Yuzhen Huang and Qian Liu and Wei Liu and Keqing He and Zejun Ma and Junxian He},
  journal={ArXiv},
  year={2025},
  volume={abs/2503.18892},
  url={https://api.semanticscholar.org/CorpusID:277940848}
}

@article{Guan2025rStarMathSL,
  title={rStar-Math: Small LLMs Can Master Math Reasoning with Self-Evolved Deep Thinking},
  author={Xinyu Guan and Li Lyna Zhang and Yifei Liu and Ning Shang and Youran Sun and Yi Zhu and Fan Yang and Mao Yang},
  journal={ArXiv},
  year={2025},
  volume={abs/2501.04519},
  url={https://api.semanticscholar.org/CorpusID:275357888}
}

@article{Wang2025ReinforcementLF,
  title={Reinforcement Learning for Reasoning in Large Language Models with One Training Example},
  author={Yiping Wang and Qing Yang and Zhiyuan Zeng and Liliang Ren and Lucas Liu and Baolin Peng and Hao Cheng and Xuehai He and Kuan Wang and Jianfeng Gao and Weizhu Chen and Shuohang Wang and Simon Shaolei Du and Yelong Shen},
  journal={ArXiv},
  year={2025},
  volume={abs/2504.20571},
  url={https://api.semanticscholar.org/CorpusID:278171513}
}

@article{Cui2025ProcessRT,
  title={Process Reinforcement through Implicit Rewards},
  author={Ganqu Cui and Lifan Yuan and Zefan Wang and Hanbin Wang and Yuchen Zhang and Wendi Li and Bingxiang He and Yuchen Fan and Tianyu Yu and Qixin Xu and Weize Chen and Jiarui Yuan and Huayu Chen and Kaiyan Zhang and Xingtai Lv and Shuo Wang and Yuan Yao and Xu Han and Hao Peng and Yu Cheng and Zhiyuan Liu and Maosong Sun and Bowen Zhou and Ning Ding},
  journal={ArXiv},
  year={2025},
  volume={abs/2502.01456},
  url={https://api.semanticscholar.org/CorpusID:276107672}
}

@article{Yu2025DAPOAO,
  title={DAPO: An Open-Source LLM Reinforcement Learning System at Scale},
  author={Qiying Yu and Zheng Zhang and Ruofei Zhu and Yufeng Yuan and Xiaochen Zuo and Yu Yue and Tiantian Fan and Gaohong Liu and Lingjun Liu and Xin Liu and Haibin Lin and Zhiqi Lin and Bole Ma and Guangming Sheng and Yuxuan Tong and Chi Zhang and Mofan Zhang and Wang Zhang and Hang Zhu and Jinhua Zhu and Jiaze Chen and Jiangjie Chen and Chengyi Wang and Honglin Yu and Weinan Dai and Yuxuan Song and Xiang Wei and Haodong Zhou and Jingjing Liu and Wei Ma and Ya-Qin Zhang and Lin Yan and Mu Qiao and Yong-Xu Wu and Mingxuan Wang},
  journal={ArXiv},
  year={2025},
  volume={abs/2503.14476},
  url={https://api.semanticscholar.org/CorpusID:277104124}
}

@software{Kydlicek_Math-Verify_Math_Verification,
author = {Kydlíček, Hynek},
license = {Apache-2.0},
title = {{Math-Verify: Math Verification Library}},
url = {https://github.com/huggingface/math-verify},
year={2025},
version = {0.6.1}
}

@article{cheng2024fullstack,
  title={FullStack Bench: Evaluating LLMs as Full Stack Coders},
  author={Cheng, Yao and Chen, Jianfeng and Chen, Jie and Chen, Li and Chen, Liyu and Chen, Wentao and Chen, Zhengyu and Geng, Shijie and Li, Aoyan and Li, Bo and others},
  journal={arXiv preprint arXiv:2412.00535},
  year={2024}
}
\bibliographystyle{icml2026}

\newpage
\appendix
\onecolumn    
\section{Training Details}
\subsection{Setup}

\begin{table*}[h]
\centering
\small
\setlength{\tabcolsep}{10pt}
\renewcommand{\arraystretch}{1.15}
\caption{Key hyperparameters used in DREAM-R.
Rollout parameters control speculative reasoning behavior,
training parameters define draft model optimization,
and reward parameters specify the SAPO objective.}
\label{tab:key_hparams}
\begin{tabular}{llll}
\toprule
\textbf{Category} & \textbf{Hyperparameter} & \textbf{Value} & \textbf{Description} \\
\midrule
Rollout & Acceptance threshold & 0.7 & Confidence threshold for accepting draft steps \\
Rollout & Max prompt length & 2048 & Maximum input context length \\
Rollout & Max response length & 8192 & Maximum generated reasoning length \\
\midrule
Training & Batch size & 64 & Number of samples per optimization step \\
Training & Learning rate & $1\times10^{-6}$ & Step size for parameter updates \\
Training & Epochs & 15 & Number of full passes over training data \\
Training & Precision & BF16 & Numeric precision for training \\
\midrule
Reward & Outcome reward weight & 1.0 & Weight on final answer correctness \\
Reward & Draft alignment reward weight & 1.0 & Weight on target verification alignment \\
Reward & Length penalty reward weight & 1.0 & Weight on discouraging long reasoning \\
\bottomrule
\end{tabular}
\end{table*}

Table~\ref{tab:key_hparams} summarizes the key hyperparameters used in DREAM-R.
For decoding, we fix the acceptance threshold to 0.7 and limit the maximum prompt
and response lengths to 2048 and 8192 tokens, respectively, ensuring stable speculative
verification while bounding the cost of long reasoning traces.

All models are trained with a global batch size of 64 and a learning rate of
$1\times10^{-6}$ using the Adam optimizer. Training is performed for 15 epochs
with BF16 precision, which provides a good balance between numerical stability
and computational efficiency.

SAPO employs a composite reward with equal weighting across its components.
Specifically, the outcome reward encourages final answer correctness, the draft
alignment reward promotes agreement with target verification, and a length penalty
discourages excessively long responses through an overlong buffer mechanism.
Unless otherwise stated, these hyperparameters are shared across all experiments.

\subsection{Training Cost}

Training a single Qwen3-VL-4B draft model with SAPO requires approximately 74 hours 
on 8$\times$NVIDIA H200 GPUs. 
The target model is queried via API during rollout and verification. 
This setup provides a practical balance between training stability and computational cost, 
enabling scalable reinforcement learning for multimodal speculative reasoning.

\section{Prompts for DREAM-R}
\label{sec:prompts}

We design two complementary prompts in DREAM-R to separate reasoning generation from verification.
The \emph{Reasoning Prompt} (Prompt~\ref{prompt:rp}) guides both the draft and target models to produce
structured reasoning, while the \emph{Scoring Prompt} (Prompt~\ref{prompt:sp}) is used exclusively by
the target model to evaluate correctness and alignment.

\newcommand{\code}[1]{\texttt{#1}}
\lstdefinestyle{wraptt}{
  basicstyle=\ttfamily\small,
  breaklines=true,
  breakatwhitespace=true,
  columns=fullflexible,
  keepspaces=true,
  backgroundcolor=\color{gray!10}
}

\begin{promptbox}{Reasoning Prompt}
\label{prompt:rp}

\small\setstretch{1.15}\RaggedRight

You are a reasoning agent responsible for generating a single coherent reasoning step
toward solving the given problem.

\textbf{Input}\\
\code{\{problem\}}\\
\code{\{image\}}, \code{\{options\}}

\textbf{Instructions}
\begin{itemize}
  \item Produce exactly one reasoning step.
  \item The step must logically follow from all previous steps.
  \item Do not generate the final answer unless explicitly required.
\end{itemize}

\textbf{Output Format}
\begin{lstlisting}[style=wraptt]
<reasoning_step>
\end{lstlisting}

\end{promptbox}

\begin{promptbox}{Scoring Prompt}
\label{prompt:sp}

\small\setstretch{1.15}\RaggedRight

You are a verification agent evaluating the correctness of the final reasoning step.

\textbf{Input}
\begin{itemize}
  \item Full problem description
  \item All previous reasoning steps
  \item The final candidate step
\end{itemize}

\textbf{Decision Rules}
\begin{itemize}
  \item Reply \code{positive} only if the step is factually correct and logically valid.
  \item Reply \code{negative} otherwise.
\end{itemize}

\textbf{Output Format}
\begin{lstlisting}[style=wraptt]
positive | negative
\end{lstlisting}

\end{promptbox}

\end{document}